\documentclass[10pt,twocolumn,letterpaper]{article}

\usepackage{iccv}
\usepackage{times}
\usepackage{graphicx}
\usepackage{amsmath}
\usepackage{amssymb}
\usepackage{multirow}
\usepackage{booktabs}
\usepackage[pagebackref=true,breaklinks=true,letterpaper=true,colorlinks,bookmarks=false]{hyperref}

\def\iccvPaperID{3877}

\iccvfinalcopy

\begin{document}

\title{Scale Matters: Temporal Scale Aggregation Network for Precise Action Localization in Untrimmed Videos}

\author{Guoqiang Gong$^1$, Liangfeng Zheng$^1$, Kun Bai$^2$, Yadong Mu$^{1,\ast}$\\
$^1$Peking University, Beijing 100871, P.R. China\\
$^2$Cloud and Smart Industries Group, Tencent China\\
{\tt\small \{1801111378,zhengliangfeng,myd\}@pku.edu.cn, kunbai@tencent.com}
}

\maketitle

\begin{abstract}
Temporal action localization is a recently-emerging task, aiming to localize video segments from untrimmed videos that contain specific actions. Despite the remarkable recent progress, most two-stage action localization methods still suffer from imprecise temporal boundaries of action proposals. This work proposes a novel integrated temporal scale aggregation network (TSA-Net). Our main insight is that ensembling convolution filters with different dilation rates can effectively enlarge the receptive field with low computational cost, which inspires us to devise multi-dilation temporal convolution (MDC) block. Furthermore, to tackle video action instances with different durations, TSA-Net consists of multiple branches of sub-networks. Each of them adopts stacked MDC blocks with different dilation parameters, accomplishing a temporal receptive field specially optimized for specific-duration actions. We follow the formulation of boundary point detection, novelly detecting three kinds of critical points (\ie, starting / mid-point / ending) and pairing them for proposal generation. Comprehensive evaluations are conducted on two challenging video benchmarks, THUMOS14 and ActivityNet-1.3. Our proposed TSA-Net demonstrates clear and consistent better performances and re-calibrates new state-of-the-art on both benchmarks. For example, our new record on THUMOS14 is $46.9\%$ while the previous best is $42.8\%$ under mAP@0.5.
\end{abstract}

\vspace{-0.1in}
\section{Introduction}
\label{sec:intro}

\renewcommand*{\thefootnote}{\fnsymbol{footnote}}
\footnotetext{$^\ast$ is the corresponding author.}
\setcounter{footnote}{0}
\renewcommand*{\thefootnote}{\arabic{footnote}}

\begin{figure}[t]
\begin{center}
\includegraphics[width=\linewidth]{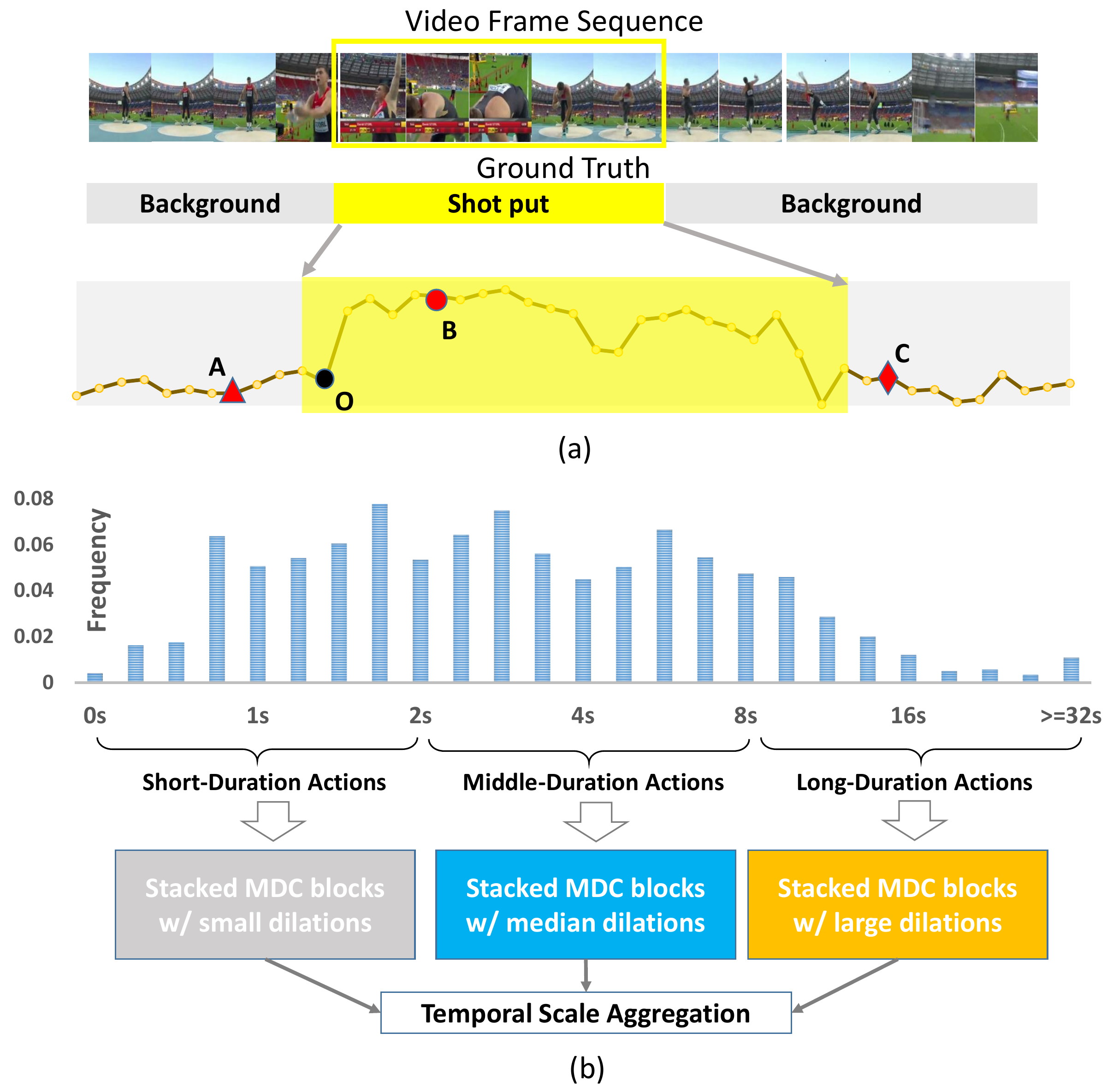}
\end{center}
   \caption{Motivation of temporal scale aggregation. Sub-figure (a) draws an untrimmed video that contains an action instance, and the corresponding actionness sequence. To faithfully detect a boundary point (such as point $O$), it is necessary to have a sufficiently large temporal receptive field that reaches points $A$, $B$, and $C$. Sub-figure (b) shows the diversity of a true action instance's time durations on THUMOS14. It is important to ensure the model can combat temporal scale variations. We propose to ensemble three-branched network with varying dilation rates, with each branch specially optimized for the specific duration. }
\label{fig:motivation}
\vspace{-0.2in}
\end{figure}

This paper addresses the task of accurately finding the temporal boundary of a video segment from an untrimmed video which instantiates specific semantic action (\eg, ``throw a frisbee" or ``parkour"), referred to as video action localization~\cite{AlwasselHG18,HeilbronLJG18,ShouCZMC17,ShouGZMC18,ShouWC16,ZhaoXWWTL17,XuDS17,corr/abs-1810-11794,ZhongLKZLL18,corr/abs-1811-07460} in the literature. This technique can serve a variety of applications in video analysis, including detecting highlights in a long video, semantic video summarization, densely captioning events in a video~\cite{KrishnaHRFN17,ZhouZCSX18,WangJ00X18} etc.
	
Given a video with $T$ frames, a naive action localization method needs to exhaustively check as many as $\mathcal{O}(T^2)$ candidate video segments, which is beyond the scope of any practitioner. Inspired by the recent advance in region CNN based image object detection~\cite{GirshickDDM14}, a majority of existing action localization methods have adopted a two-step strategy. The first step performs a quick screening over all candidates and only most potential video segments remain for further inspection, called \emph{video action proposals}. The second step further conducts a fine-grained evaluation on proposals. An ideal set of action proposals shall have both high recall rate and temporal overlapping with true action instances.

Tremendous recent endeavors have been devoted to improving the quality of video action proposals, which can be roughly casted into three main thrusts. The first family of methods (such as S-CNN~\cite{ShouWC16} or TURN~\cite{GaoYSCN17}) generates video segments using sliding windows of pre-determined length, which are typically followed by some segment-level ranker. The second type of methods utilizes temporal actionness grouping. They estimate actionness for each frame or video snippet and group several consecutive confident frames into an action proposal. Some recent works, exemplified by CTAP~\cite{GaoCN18}, combine sliding windows and actionness estimation for generating better proposals. The third thrust learns to detect exact temporal boundaries through frame-wise estimation of staring / ending points. In practice, adjacent detected boundary points will be paired to form valid proposals, as done in the recently-proposed boundary-sensitive network (BSN)~\cite{LinZSWY18}.

\begin{figure*}[t]
\begin{center}
\includegraphics[width=0.85\linewidth]{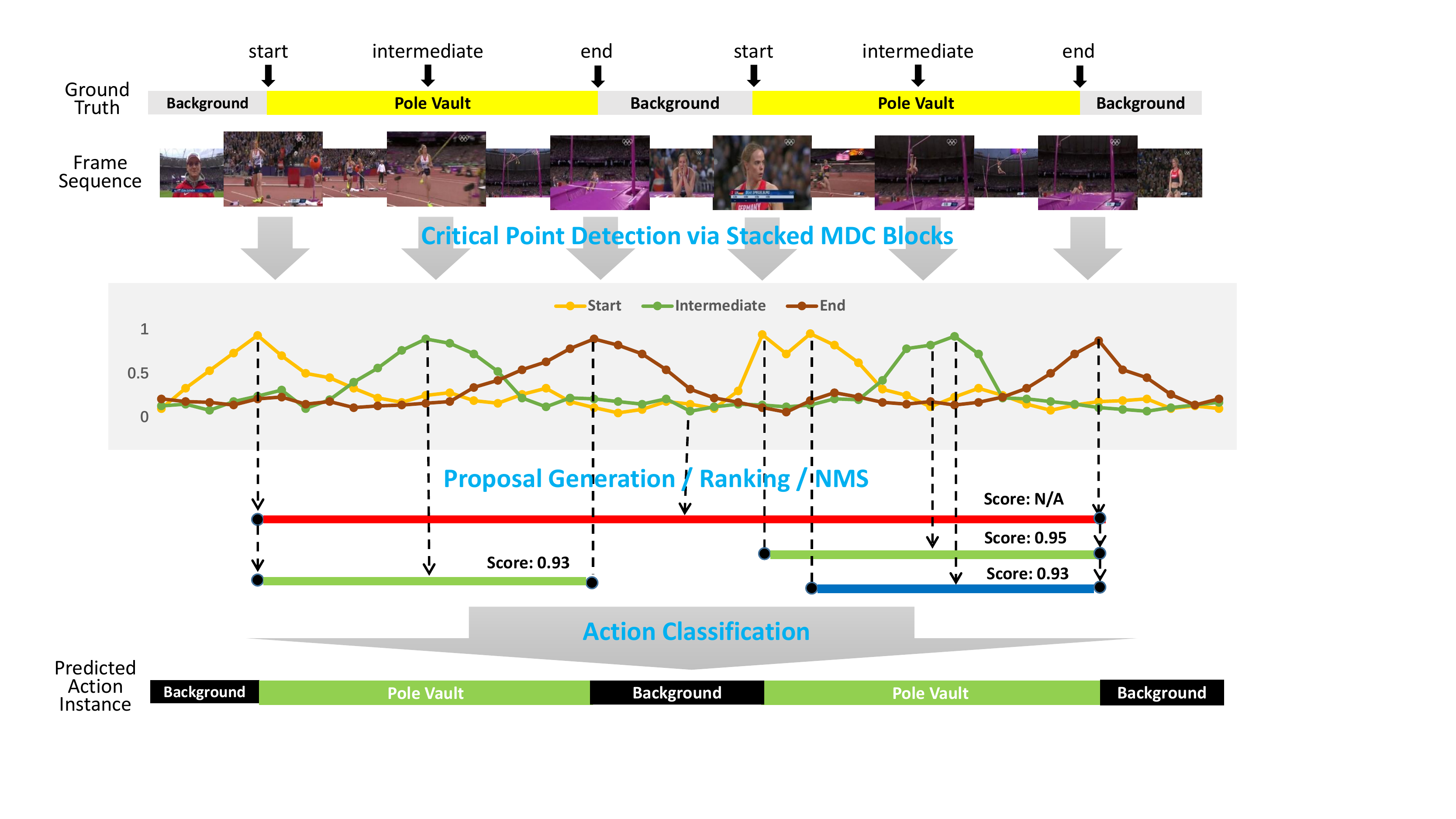}
\end{center}
   \caption{\small Pipeline of our proposed temporal scale aggregation (TSA) network. See Section~\ref{sec:tsa} for more details.}
\label{fig:overview}
\vspace{-0.15in}
\end{figure*}

The motivating observation for our work is illustrated in Figure~\ref{fig:motivation}(b). As seen, durations of true action instances can diversely distribute, usually varying from few seconds to several minutes. Most of previous methods utilize a procrustean way to tackle the temporal scale issue. For temporal sliding window or actionness grouping based methods (\eg,~\cite{ShouWC16,ZhaoXWWTL17}), a popular treatment is to build a video pyramid via temporal sub-sampling and extract same-sized video segments from all pyramid levels. An action instance will be detected at specific temporal scales where its re-sized duration is amenable for detection, ignored at other scales. However, the video pyramid significantly complicates learning an effective model and often brings inferior performance than other methods. On the other hand, boundary point based methods, such as BSN~\cite{LinZSWY18}, typically ignore the scale issue and use fixed convolutional receptive field for all action instances.

This work proposes \emph{temporal scale aggregation} network (TSA-Net) for video action localization. The key differentiator from other competing methods is novelly using multi-dilation temporal convolutions for seriously addressing the temporal scale issue. Figure~\ref{fig:motivation} illustrates our key insight. Investigating temporal context is critical when confidently localizing some boundary points. However, as the duration of an action instance varies, finding a one-for-all temporal receptive field is arguably not possible. As a natural solution, we propose to ensemble a set of parameterized temporal convolutional building blocks. Each of them has different receptive field that is most effective at specific temporal scale. The responses of all temporal convolutions are fused to more reliably estimate a boundary point. The main technical contributions of this work are summarized as below:

1. We propose an action localization model TSA-Net, which falls into a two-step framework (proposal generation + classification). Unlike previous works, TSA-Net concurrently considers many temporal scales when estimating the boundary points. Specifically, we devise a \emph{multi-dilation temporal convolution} (MDC) block as the core component of TSA-Net. To our best knowledge, it has been seldom explored in video action localization to tackling plenary temporal scales by efficiently manipulating convolutional dilation rates.

2. We design TSA-Net as a marriage of boundary detection and actionness grouping. It simultaneously detects three kinds of points: starting, ending and the mid-point of an instance. The mid-point implicitly encodes the actionness information of a proposal. A starting / ending pair can only be useful together with a confident mid-point. This way it effectively avoids the low-accuracy problem in previous boundary point based methods (\eg, BSN~\cite{LinZSWY18}).

3. Comprehensive evaluations are conducted on two challenging benchmarks: THUMOS14 and ActivityNet-1.3. TSA-Net outperforms all competitors by large margins, re-calibrating  state-of-the-art performance on both benchmarks. For example, our new record on THUMOS14 is $46.9\%$ while the previous best is $42.8\%$ for mAP@0.5.

\section{Related Work}

The goal of temporal action localization is precisely detecting video segments with specific actions from untrimmed videos. A majority of action localization methods~\cite{ShouWC16,ZhaoXWWTL17,XuDS17,DaiSZDC17,ChaoVSRDS18} adopt a two-stage computational pipeline, strongly inspired by the success of two-stage image object detection~\cite{GirshickDDM14,Girshick15,RenHGS15,KuoHM15,LinDGHHB17}. The first step does the job of proposal generation, which filters out the most irrelevant video segments. The second step further learns to classify action proposals and de-limit their exact temporal boundaries. Besides the two-stage methods, there also exist other methods which adopt a framework of reinforcement learning~\cite{YeungRMF16} or single-shot detection~\cite{BuchEGFN17,LinZS17MM,ZhangDWW18} inspired by their counterparts in image object detection (\eg, YOLO~\cite{RedmonDGF16} and SSD~\cite{LiuAESRFB16}).

As briefly summarized in Section~\ref{sec:intro}, popular proposal-generating schemes can be roughly categorized as sliding windows~\cite{EscorciaHNG16,HeilbronNG16,ShouWC16,BuchESGN17,GaoYSCN17}, temporal action grouping~\cite{ZhaoXWWTL17}, or boundary point detection~\cite{LinZSWY18} etc. The evaluation of an action localization model takes intersection-over-union (IoU) between an action proposal and its closest true instance into account. Proposals below some pre-defined IoU threshold are regarded as false positives. State differently, action proposals partially overlapping with some ground-truth video segments are not favored. It thus spurs researchers to develop models that generate action proposals with more accurate temporal boundaries. Two caveats for obtaining good temporal boundary are doing the job at finer temporal scales, as shown by Convolution-De-Convolution(CDC)~\cite{ShouCZMC17}, and fully utilizing temporal contextual information, as demonstrated by SSN~\cite{ZhaoXWWTL17} that decomposes an action proposal into starting-course-ending phases or BSN~\cite{LinZSWY18} that conducts stacked temporal convolutions.

Dilated convolutions have been widely adopted in many tasks for capturing long-range dependency, including video salient object detection~\cite{SongWZSL18}, image classification~\cite{YuK15,YuKF17}, semantic image segmentation~\cite{WeiXSJFH18,MehtaRCSH18}, crowd analysis~\cite{LiZC18}, action recognition~\cite{XuYZWLJ18} or segmentation~\cite{LeaFVRH17} etc. The work in~\cite{ChenPKMY18} pioneered the endeavor that uses a few convolutions with different dilated rates for aggregating multi-scale context.
Though dilated convolutions have been previously used in action localization~\cite{ChaoVSRDS18}, it is mainly for enlarging temporal receptive field. Exploring temporal context across scales is still insufficiently studied in action localization. To our best knowledge TSA-Net is the first of its kind.

\vspace{-0.08in}
\section{Temporal Scale Aggregation Network}
\label{sec:tsa}

This section details our proposed TSA-Net, whose computational pipeline is depicted in Figure~\ref{fig:overview}. We follow a similar design inspired by~\cite{LinZSWY18,ChaoVSRDS18}, but have essential differences in concurrently tackling multiple temporal scales. The sub-sections below will focus on our novelty and omit trivial technical details shared with other methods.

\subsection{Video Feature Extraction}
\label{sec:feature}

Let $V$ be an untrimmed video of $T$ frames. In the case of over-long videos, we proceed to extract video features with a regular frame interval $ \sigma $ in order to reduce the computation cost, resulting in $T \leftarrow T / \sigma $ video snippets. Without causing much confusion, we stick to misuse the variable $T$ for representing frame or snippet count. Following the common practice in previous works, two-stream video features~\cite{SimonyanZ14} are extracted for each snippet. For ablatively studying the effect of different video features, P3D~\cite{QiuYM17} or C3D~\cite{TranBFTP15} features are also adopted. Let $F= [f_1,f_2,\ldots,f_{ T}]$ be the feature sequence. More details regarding $F$ are deferred to Section~\ref{sec:exp_setup}.

\subsection{Critical Point Detection}
\label{sec:point}

The goal of this step is to estimate each video snippet's probability of corresponding to a starting / intermediate / ending moment of an action instance.

\begin{figure}[t]
\begin{center}
\includegraphics[width=\linewidth]{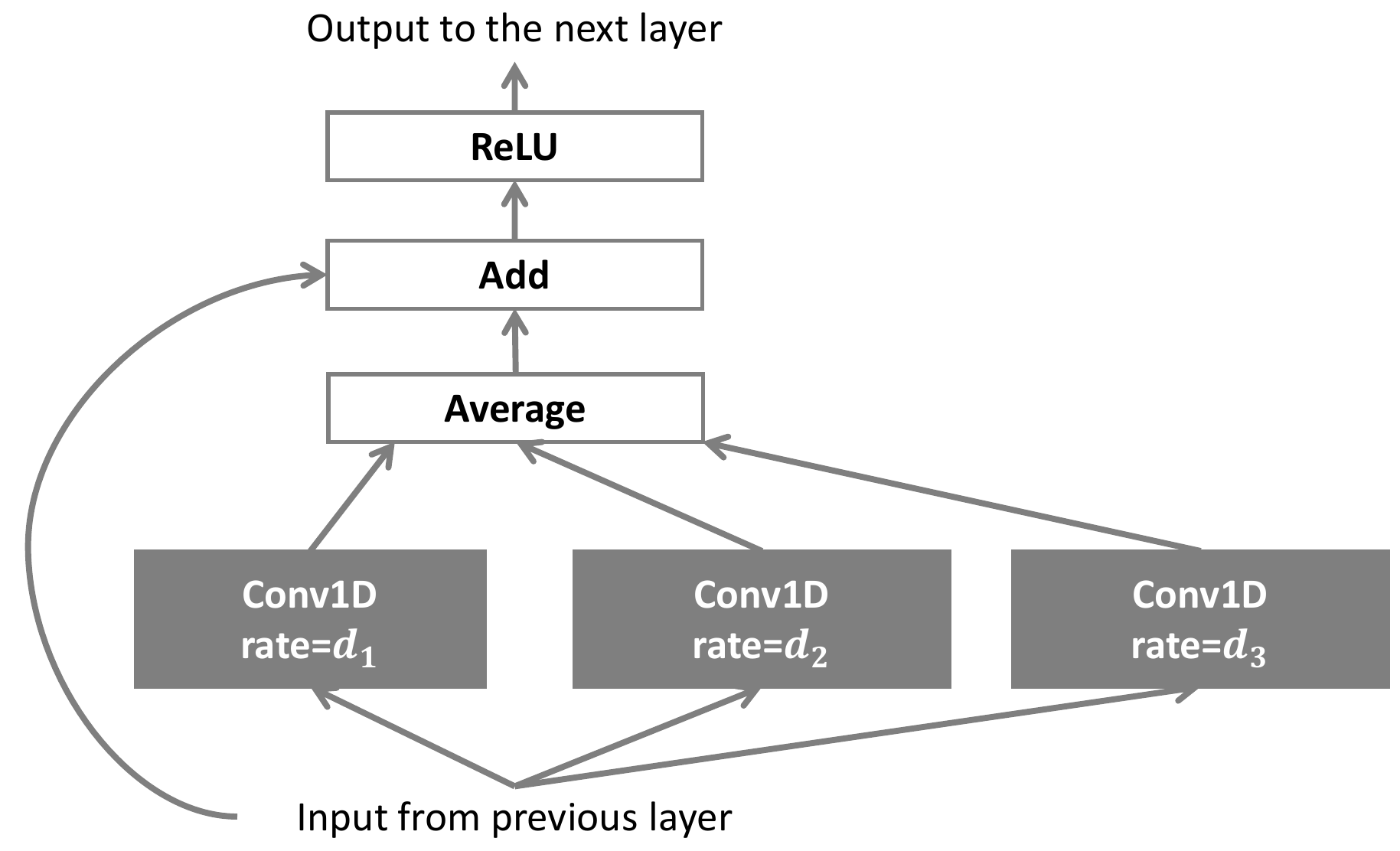}
\end{center}
   \caption{\small Multi-dilation temporal convolution (MDC) block. $d1$, $d2$ and $d3$ are the dilation rates of temporal 1D convolutions.}
\label{fig:block}
\vspace{-0.15in}
\end{figure}

\noindent \textbf{Multi-Dilation Temporal Convolution (MDC) Block}: Figure~\ref{fig:motivation} emphasizes the significance of choosing proper temporal receptive field when trying to detect a boundary point or mid-point (hereafter we call it \emph{critical point}). Consecutive video frames tend to be visually correlated, and investigating video frames sufficiently far away (via temporal 1-D convolution with video features $F$) is important for finding critical points. To enlarge the receptive field under budgeted computation, we devise a convolutional unit that ensembles multiple dilation rates. The architecture is illustrated in Figure~\ref{fig:block}. All dilated temporal 1-D convolutions have the same kernel size, yet with a typical choice $d_1 < d_2 < d_3$ that defines increasingly larger receptive fields. Denote it with the notation MDC-$(d_1,d_2,d_3)$. A dilation rate of 1 boils down to a normal convolution. The outputs from all dilated convolutions are simply averaged, returning fused contextual information. Note that a skip connection is inserted after the average operation, such that the dilated convolutions are reinforced to focus on learning the residual.

\noindent \textbf{Multi-branch Stacked MDCs}: The duration of video action instances widely varies, typically ranging from 1/10 seconds to a few minutes. For actions of different durations, the required scales of temporal context information are different. To localize short-duration action instances, over-large receptive field of temporal convolutions can harmfully bring irrelevant information that distracts the parameter optimization. While for long action instances, a small temporal receptive field may miss some key discriminative information, such as being unable to reach all points $A$, $B$ and $C$ when judging $O$ as in Figure~\ref{fig:motivation}(a).

\begin{figure}[t]
\begin{center}
\includegraphics[width=0.9\linewidth]{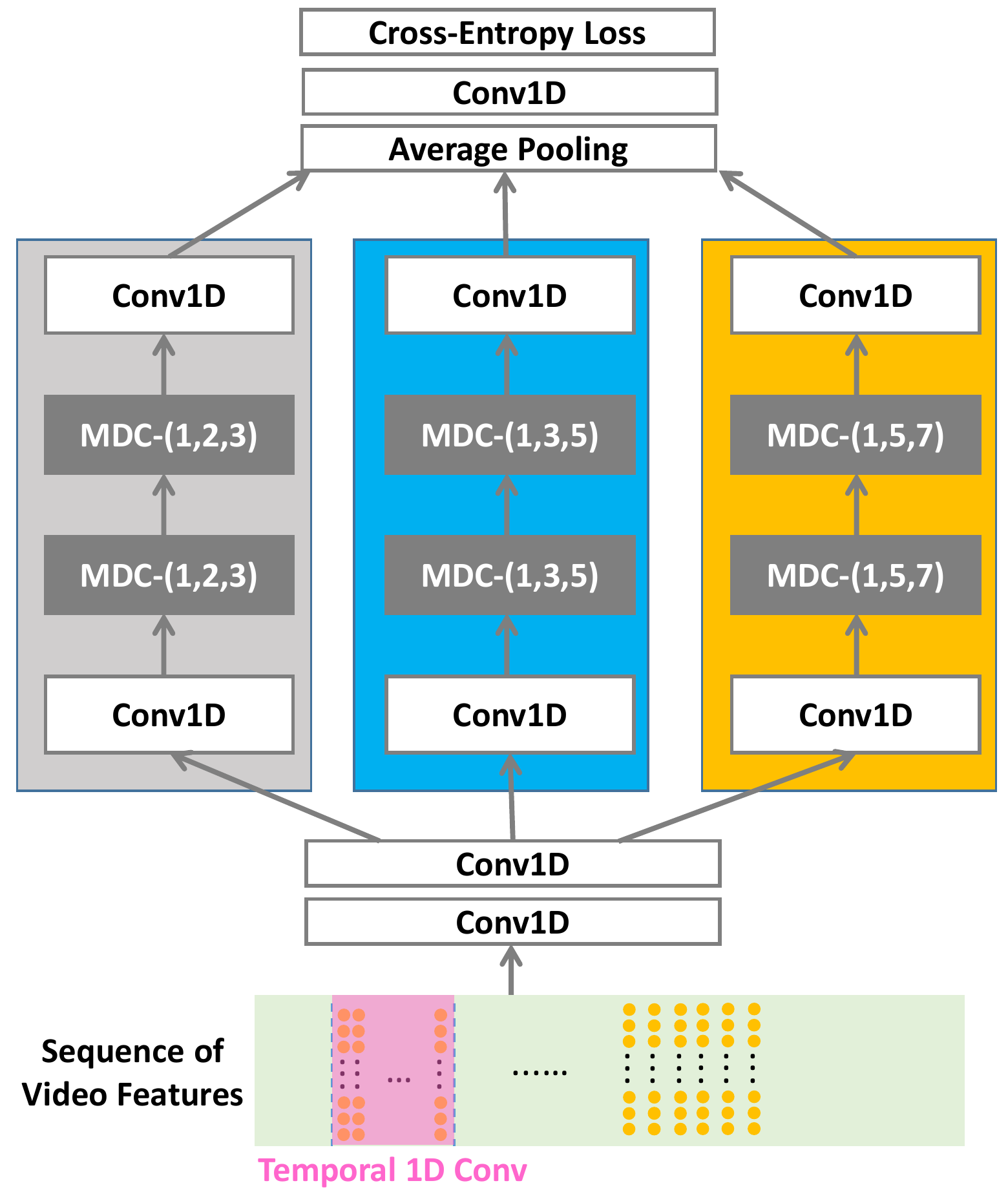}
\end{center}
   \caption{\small The architecture of multi-branch stacked MDC blocks which calculates critical point's probability sequences $P^{(s)}, P^{(i)}, P^{(e)} \in [0,1]^T$. Each branch is designed for tackling videos of specific durations. We omit all ReLU and sigmoid activation layers for saving space.}
\label{fig:network}
\vspace{-0.2in}
\end{figure}

Unfortunately, it is arguably impossible for finding a one-for-all temporal receptive field. We propose to novelly ensemble multiple network branches. Each of these branches contains stacked MDC blocks parameterized with different dilation rates. State differently, we customize each branch such that it captures videos of specific durations. The network architecture is shown in Figure~\ref{fig:network}. As seen, the input video snippet features first go through several shared 1-D temporal convolutions for lightly exchanging among adjacent frames. Three branches of stacked MDC blocks are followed. The dilation rates directly control the extent of receptive field. For example, a 2-stack MDC-$(1,2,3)$ (\ie, the leftmost branch) or MDC-$(1,5,7)$ (\ie, the rightmost) implies a receptive fields of 13 or 29 respectively. One can flexibly tailor the dilation parameters or branch count to fit the videos under inspection. Outputs from all branches are pooled for further processing (we use average pooling).

\noindent \textbf{Objective of Critical Point Detection}: Given an untrimmed video $V$ of $T$ frames (or snippets) and the corresponding features $F= [f_1,f_2,\ldots,f_T]$, each starting / ending point pair $(t^{(s)}, t^{(e)})$ temporally de-limits an action instance in this video and also implies an intermediate point $t^{(i)} = (t^{(s)}+ t^{(e)})/2$. Let $\Gamma = \{ \gamma_k = (t^{(s)}_k, t^{(e)}_k, t^{(i)}_k), k=1 \ldots K \} $ be a collection of $K$ annotations in the video $V$.

Motivated by recent advance in image translation~\cite{IsolaZZE17}, we formulate critical point detection as a sequence-to-sequence learning problem. Specifically, let us introduce three notations $Y^{(s)}, Y^{(i)}, Y^{(e)} \in \{0,1\}^T$ to denote the ground truth of starting / intermediate / ending points, respectively. A direct treatment is to set all $t^{(s)}_k, k=1\ldots K$ in $Y^{(s)}$ to 1, otherwise 0. However, since the critical point is highly sparse, such treatment leads to heavy imbalance between positive and negative labels. To mitigate this issue, we inflate each annotated critical point, such as expanding point $t^{(s)}_k$ to a region $[t^{(s)}_k - \delta \cdot L_k, t^{(s)}_k + \delta \cdot L_k]$, where $L_k=t^{(e)}_k-t^{(s)}_k$ and $\delta$ is a hyper-parameter (we empirically set to 0.1 in all experiments). All time locations in the expanded region is set to be positive. Likewise, $Y^{(i)}, Y^{(e)}$ are calculated in a similar protocol.

In Figure~\ref{fig:network}, the outputted feature maps from three branches are aggregated via average pooling. After going through some Conv1D and sigmoid layers, the network eventually renders three probability sequences $P^{(s)}, P^{(i)}, P^{(e)} \in [0,1]^T$, describing our estimation for a point belonging to starting / intermediate / ending respectively. The pooling-based aggregation ensures a salient probabilistic response, once any branch of stacked MDCs captures the mid-point or boundary-like pattern around this critical point.

The overall objective to be minimized is defined as the sum over three kinds of critical points, \ie, $\mathcal{J} = \mathcal{J}^{(s)} + \mathcal{J}^{(i)} + \mathcal{J}^{(e)}$, where each term on the right hand side adopts a cross-entropy loss. For example,
\begin{eqnarray}
\mathcal{J}^{(s)} &=&  \sum_{i=1 \ldots T} \Large[  Y^{(s)}(i) \cdot \log( P^{(s)}(i) ) \nonumber \\
&& + (1-Y^{(s)}(i)) \cdot \log( (1-P^{(s)}(i)) ) \Large].
\end{eqnarray}

\subsection{Proposal Generation and Ranking}
\label{sec:proposal}

The set of action proposals for a video are obtained via sequentially conducting the operations below:

\noindent \textbf{Critical Point Selection and Pairing}: The probability sequences $P^{(s)}, P^{(i)}, P^{(e)} \in [0,1]^T$ are designed to indicate how likely a point is a critical point. Therefore one can find critical points by simply thresholding the probability values, say 0.9 as we adopt in all experiments. However, this strategy often misses many true critical points. In order to elevate the recall rate, we also find local maxima by comparing a point with its temporal neighbors and mark them as critical points. We perform above operations on both $P^{(s)}$ and $P^{(e)}$, obtaining candidate starting / ending point sets $C^{(s)} = \{c^{(s)}\}$, $C^{(e)} = \{c^{(e)}\}$ respectively.

Next, any two points $c^{(s)} \in C^{(s)}$ and $c^{(e)} \in C^{(e)}$ will be paired to generate a proposal when they satisfy the conditions below:

1. The distance between $c^{(s)}$ and $c^{(e)}$ is within $[d_{min},d_{max}]$,  where $[d_{min},d_{max}]$ are the smallest and largest durations estimated from annotated action instances in the training set. This way enforces strong prior on a proposal's duration and can filter out many false pairings.

2. Let $c^{(i)} = (c^{(s)} + c^{(e)}) / 2$ be the mid-point and $P^{(i)}(c^{(i)})$ be the corresponding probability as an intermediate critical point. A low value of $P^{(i)}(c^{(i)})$, as exemplified by the proposal highlighted in red in Figure~\ref{fig:overview}, implies that $c^{(s)}, c^{(e)}$ may indeed come from different action instances and the pair shall be abandoned.

\noindent \textbf{Bayesian Proposal Ranking}: The proposal set generated by previous step tends to be noisy. A follow-up step of proposal ranking can effectively remove many false negatives. In order to score an arbitrary proposal defined by boundary points $c^{(s)}, c^{(e)}$, we adopt a Bayesian formulation as below:
\begin{equation}
P \left( \overline{c^{(s)}_i c^{(e)}_i} \right) = P^{(s)}(c^{(s)}_i) \cdot P^{(e)}(c^{(e)}_i) \cdot \phi(c^{(s)}_i, c^{(e)}_i), \label{eqn:bayesian}
\end{equation}
where $P^{(s)}, P^{(e)}$ are afore-mentioned point-wise probabilities of being a critical point. $\phi(c^{(s)}_i, c^{(e)}_i)~:~\mathbb{R}^+ \times \mathbb{R}^+ \mapsto [0,1]$ represents some compatibility function to be learned.

To learn $\phi(c^{(s)}_i, c^{(e)}_i)$, let us first define a feature representation for the segment $\overline{c^{(s)}_i c^{(e)}_i}$. Inspired by recently-proposed temporally-structured segment networks~\cite{ZhaoXWWTL17}, we extend $\overline{c^{(s)}_i c^{(e)}_i}$ outwards (in practice this segment is resized to $1.4 \times (c^{(e)}-c^{(s)}) $) to include more non-action background information. 32 scalar values are uniformly sampled from the resized segment by reading $P^{(s)}$, $P^{(i)}$ and $P^{(e)}$ respectively. After concatenation and vectorization, this forms a 96-dimensional representation for $\overline{c^{(s)}_i c^{(e)}_i}$, which is then fed into a small network FC($96 \rightarrow 96$) $\Rightarrow$ ReLU $\Rightarrow$ FC($96 \rightarrow 48$) $\Rightarrow$ ReLU $\Rightarrow$ FC ($48 \rightarrow 1$) $\Rightarrow$ Sigmoid. The last sigmoid layer outputs a probabilistic value, which is exactly $\phi(c^{(s)}_i, c^{(e)}_i)$.

To enforce the learned $\phi(c^{(s)}_i, c^{(e)}_i)$ to be informative, we borrow the idea in~\cite{LinZSWY18} and solve:
\begin{equation}
\min \sum _i \ell \left( \phi(c^{(s)}_i, c^{(e)}_i), \sup_{\gamma_k \in \Gamma} IoU(\overline{c^{(s)}_i c^{(e)}_i}, \gamma_k) \right)  ,
\end{equation}
where the sum is calculated over all potential proposals. We choose $\ell(\cdot)$ to the smooth ${ L }_{ 1 }$ loss~\cite{Girshick15}. $IoU(\cdot, \cdot)$ is an intersection-over-union operator between two 1-D segments.

\noindent \textbf{Redundancy Removal}: Learning to calculate $P \left( \overline{c^{(s)}_i c^{(e)}_i} \right)$ enables us to conduct non-maximum suppression (NMS) to remove redundant proposals. We experiment with either naive greedy NMS to soft-NMS \cite{BodlaSCD17}, depending on which the competing method under comparison uses. An example of removed proposal is highlighted in blue color in Figure~\ref{fig:overview}.

\subsection{Action Classification}
\label{sec:classification}

The last step of two-stage video action localization is feeding proposals into an action classifier. It categories the proposal to one of many pre-defined action classes, or the null class. Since the major scope of this work is about a novel scheme for proposal generation, we directly adopt action classifiers widely used in previous works. This eases more focused comparisons with other action localization methods. Specifically, we use UntrimmedNet (UNet)~\cite{WangXLG17} or SCNN-classifier (SCNN-cls)~\cite{ShouWC16} on THUMOS14, and another action classification model~\cite{zhao2017cuhk} on ActivityNet-1.3. Following common practice, the classification only involves 200 most confident proposals per video on THUMOS14 and top-100 proposals per video on ActivityNet-1.3.

\section{Evaluations}

\subsection{Dataset Preparation and Implementation}
\label{sec:exp_setup}

We experiment with two large-scale video benchmarks widely used for video action localization. \textbf{THUMOS14}~\cite{idrees2017thumos} contains videos from 20 sports action classes. There are 200 and 212 temporally-annotated videos in validation and testing sets respectively. On average, each video contains more than 15 action instances. Following the settings in previous works, we use 200 untrimmed videos in the validation set to train our model and evaluate on the test set. \textbf{ActivityNet-1.3}~\cite{HeilbronEGN15} consists of 19,994 videos with 200 classes annotated. Each video contains $\sim1.5$ activity instances. The entire data is divided into training, validation and testing sets using the ratio of 2:1:1.

We extract two sets of heterogeneous video features in order to investigate how different features affect the final performance, including: 1) \textbf{two-stream features}: The specific implementation of two-stream model in~\cite{XiongWWZSLL0GT16} is used, which is comprised of BN-Inception for motion and ResNet for visual appearance. We adopt the version pre-trained on ActivityNet-1.3 in~\cite{XiongWWZSLL0GT16}. For both BN-Inception and ResNet, we decapitate each net by discarding the final classifier layer, and vectorize the output of last fully-connected layer. This brings exactly both 200-D vectors as motion / appearance features respectively. They are further concatenated to form a 400-D snippet-level feature; 2) (2) \textbf{P3D~\cite{QiuYM17} or C3D~\cite{TranBFTP15} features}:Two separate P3D models are trained from consecutive frames and flows on Kinetics~\cite{CarreiraZ17} respectively. The dimensions of appearance and motion feature are both 2048-dimensional. Feature concatenation makes another 4096-D snipped-level features. The C3D model are pre-trained on the UCF-101 dataset~\cite{abs-1212-0402}. To reduce computational complexity, each video snippet is sampled for every 5 frames on THUMOS14 and 16 on ActivityNet-1.3. The parameters of 5 or 16 are chosen to well balance complexity and video information fidelity. Evaluations on THUMOS14 are performed using both two-stream and P3D / C3D features. While on ActivityNet-1.3, only two-stream feature is adapted for comparing different methods.

Our proposed TSA-Net is fully implemented in Tensorflow. On both datasets, multi-branch stacked MDC blocks in Figure~\ref{fig:network} and the sub-network for scoring $P \left( \overline{c^{(s)}_i c^{(e)}_i} \right)$ are sequentially optimized, rather than in an end-to-end fashion. Specifically, the former is trained with a batch size of 16. The learning rate of the first 10 epochs is set to 0.001, then 0.0001 for all following epochs. The whole optimization will terminate before a maximum of 20 epochs in order to avoid over-fitting. The latter is trained with a batch size of 256 and using the same scheme of learning rates.

\subsection{Evaluation Protocol}

\noindent \textbf{Action Proposal Generation}: \emph{Average recall} (AR) at different IoU is typically used as the evaluation metric. Following common practice, we use an IoU threshold set from 0.5 to 1.0 with a stride of 0.05 on THUMOS-14, and 0.5 to 0.95 with a stride of 0.05 on ActivityNet-1.3. We calculate AR with different \emph{average number} (AN) of proposals per video to evaluate the relationship between recall and proposal number. On ActivityNet-1.3, we also use the area under the AR-AN curve (AUC) as metrics, where AN varies from 0 to 100.

\noindent \textbf{Action Classification}: We adopt standard \emph{mean average precision} (mAP) metric. A prediction is regarded as positive only when the predicted action category is correct and meanwhile IoU with ground truth is above some pre-defined threshold. We report mAP at different IoU thresholds. On THUMOS14, the IoU thresholds are {0.3, 0.4, 0.5, 0.6, 0.7}. On ActivityNet-1.3, the IoU thresholds are {0.5, 0.75, 0.95}. The average mAP with IoU thresholds [0.5:0.05:0.95] is used to compare different methods.

\begin{table}[t]
\centering
\begin{scriptsize}
\begin{tabular}{clccc}
  \toprule[0.8pt]
  \multirow{2}{*}{Feature} & \multirow{2}{*}{Method} & \multicolumn{3}{c}{{AR@AN}} \\
  	& & {@50} & {@100} & {@200}  \\
  \midrule[0.5pt]
    {C3D} & {DAPs~\cite{EscorciaHNG16}} & 13.56 & 23.83	& 33.96  \\
    {C3D} & {SCNN-prop~\cite{ShouWC16}} & 17.22 & 26.17& 37.01  \\
    {C3D} & {SST~\cite{BuchESGN17}} & 19.90& 28.36 & 37.90  \\
    {C3D} & {TURN~\cite{GaoYSCN17}} & 19.63& 27.96 & 38.34  \\
    {C3D} & {BSN + Greedy-NMS~\cite{LinZSWY18}} & 27.19& 35.38 & 43.61  \\
    {C3D} & {BSN + Soft-NMS~\cite{LinZSWY18}} & 29.58& 37.38 & 45.55  \\
    {C3D} & {Ours + Greedy-NMS~} & 31.88& 40.19 & 46.28  \\
    {C3D} & {Ours + Soft-NMS~} & \textbf{34.00} & \textbf{41.11} & \textbf{47.51}  \\
  \midrule[0.5pt]
    {Flow} & {TURN~\cite{GaoYSCN17}} & 21.86	& 31.89	& 43.02  \\
    {2-Stream} & {TAG~\cite{XiongZWLT17}} & 18.55	& 29.00 & 39.41  \\
    {2-Stream} & {CTAP~\cite{GaoCN18}} & 32.49	& 42.61	& 51.97 \\
    {2-Stream} & {BSN + Greedy-NMS~\cite{LinZSWY18}} & 35.41	& 43.55	& 52.23 \\
    {2-Stream} & {BSN + Soft-NMS~\cite{LinZSWY18}} & 37.46	& 46.06	& 53.21 \\
    {2-Stream} & {Ours + Greedy-NMS} & 41.40 & 48.70  & 53.57  \\
    {2-Stream} & {Ours + Soft-NMS} & \textbf{42.83} & \textbf{49.61} & \textbf{54.52}  \\
  \midrule[0.5pt]
    {P3D} & {Ours + Greedy-NMS} & 45.84	& 52.29	& 56.17  \\
    {P3D} & {Ours + Soft-NMS} & \textbf{46.77}	 & \textbf{53.15} & \textbf{57.68}  \\
  \bottomrule[0.8pt]
\end{tabular}
\end{scriptsize}
\caption{Comparisons in terms of AR@AN on THUMOS14.}
\label{table:thumos_prop}
\vspace{-0.1in}
\end{table}

\subsection{Comparisons for Proposal Generation}

Table~\ref{table:thumos_prop} summarizes all comparisons conducted on the test set of THUMOS14. Regarding proposal generation, our proposed TSA-Net consistently outperforms other methods when AN ranges from 50 to 200. Specifically, for AR@50, our method significantly improves the performance from the previous record 37.46\% in the literature to 42.83\% using identical two-stream features. Using stronger P3D features can further elevate the AR@AN scores.

\begin{figure}[t]
\begin{center}
\includegraphics[width=0.85\linewidth,height=1.8in]{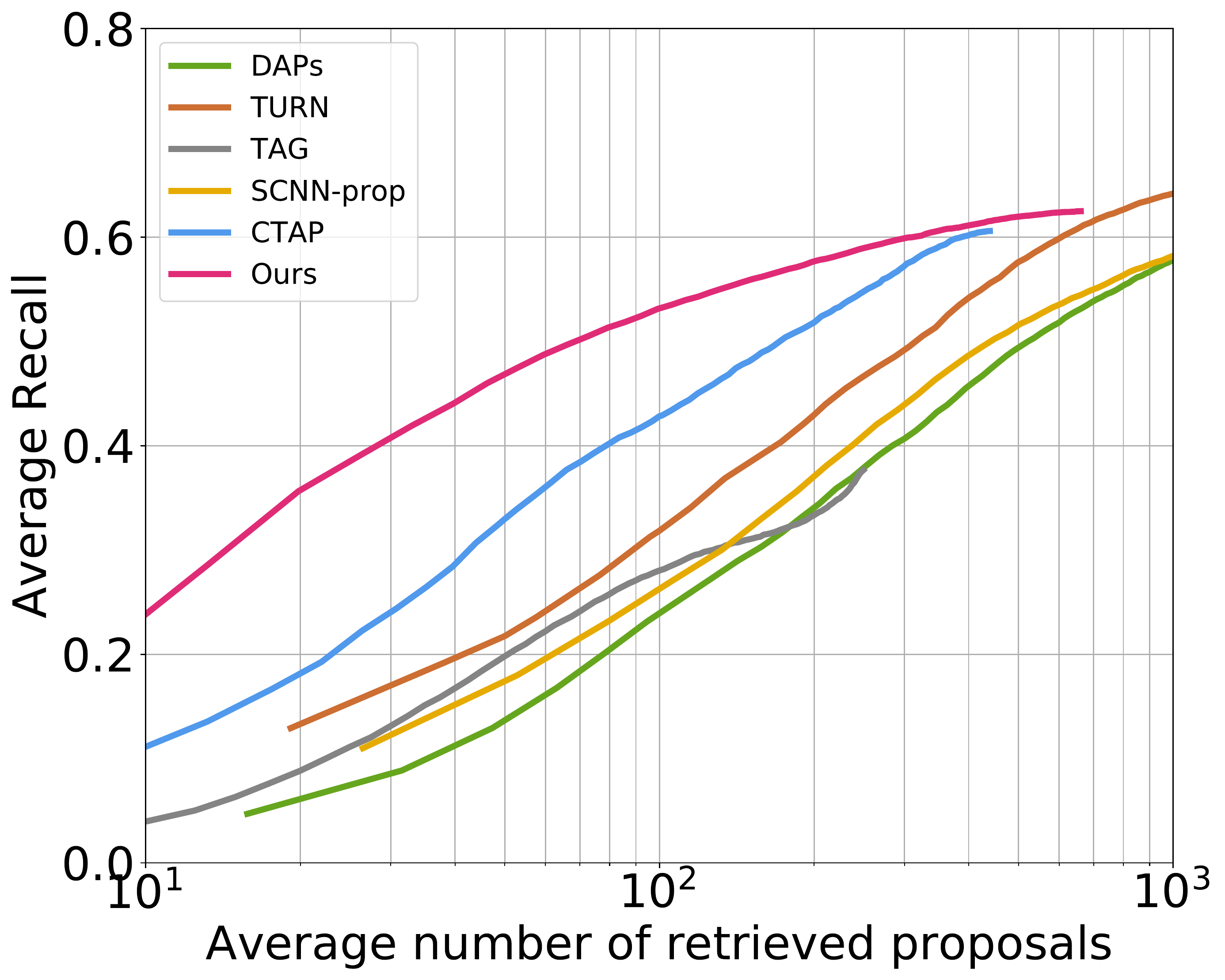}
\end{center}
   \caption{AR-AN curves of different methods on THUMOS14 test set.}
\label{fig:res_avg_recall}
\vspace{-0.15in}
\end{figure}

\begin{figure}[t]
\begin{center}
\includegraphics[width=0.85\linewidth,height=1.8in]{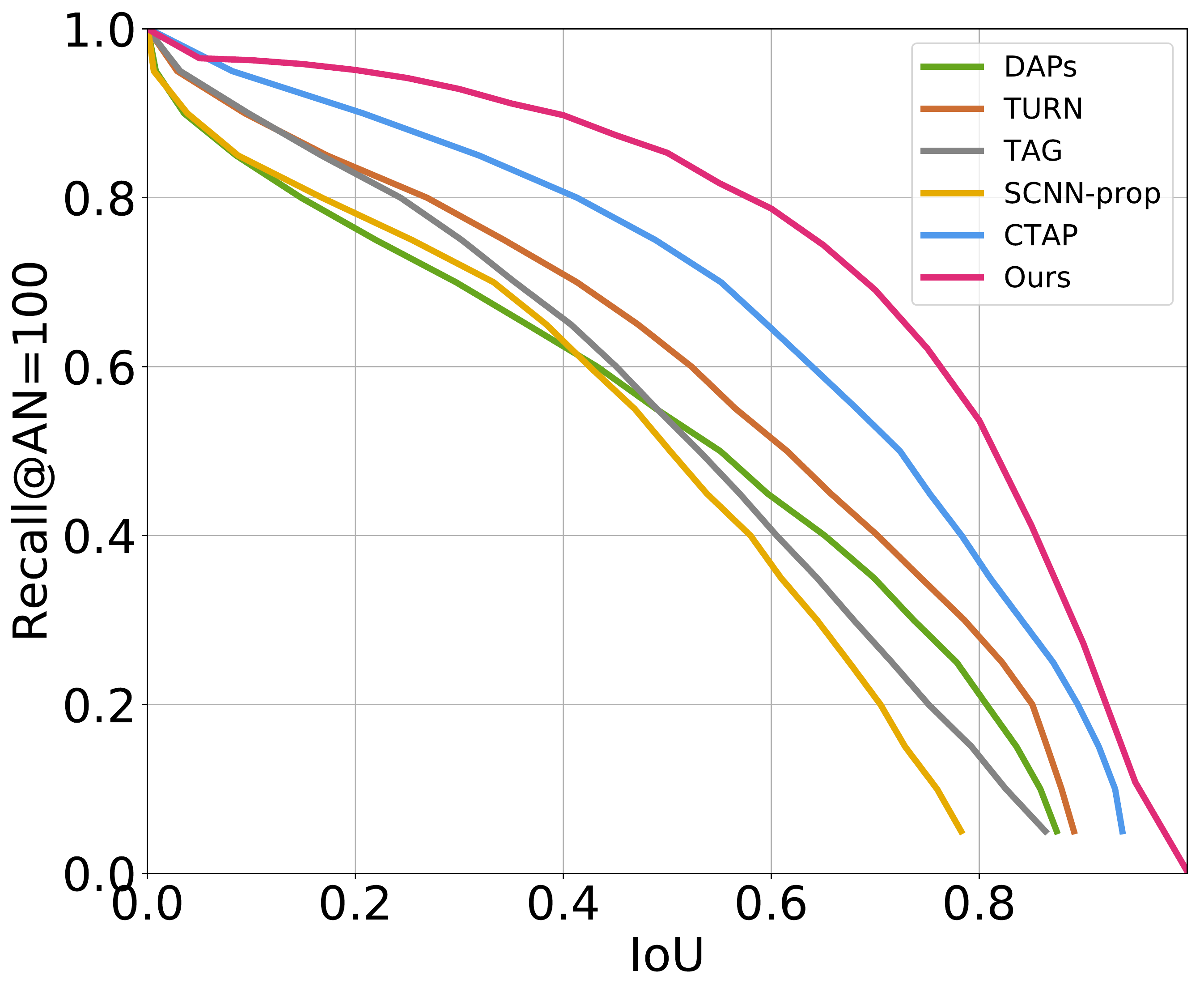}
\end{center}
   \caption{Recall@AN=100 curves of our method and previous state-of-the-art methods on THUMOS14 test set.}
\label{fig:res_recall100}
\vspace{-0.06in}
\end{figure}

Figure~\ref{fig:res_avg_recall} illustrates the AR-AN curves of different methods on THUMOS14. Clearly, our method achieves better AR in the low AN region, suggesting strong ability to generate top-ranked proposals with higher quality. To further demonstrate that the proposals generated by our method have higher IoU with ground truth, we calculate the recalls at multiple IoU thresholds using 100 proposals per video. Figure~\ref{fig:res_recall100} consistently shows the superior recall rates of our method when IoU ranges from 0 to 1.

\begin{table}[t]
\centering
\begin{small}
\begin{tabular}{lccc}
  \toprule[0.8pt]
    {Proposal Method} & {AUC(val)} & {AUC(test)} & {AR@100(val)}  \\
  \midrule[0.5pt]
    {TCN~\cite{DaiSZDC17}} & 59.58	& 61.56	& -  \\
    {Prop-SSAD~\cite{LinZS17}} & 64.80	& 64.40 & 73.01  \\
    {CTAP~\cite{GaoCN18}} & 65.72& -	& 73.17 \\
    {BSN~\cite{LinZSWY18}} & 66.17	& 66.26	& 74.16 \\
  \midrule[0.5pt]
    {Ours} & \textbf{66.82} & \textbf{67.01}  & \textbf{74.82}  \\
  \bottomrule[0.8pt]
\end{tabular}
\end{small}
\caption{\small  Performance comparisons with TCN, Prop-SSAD, CTAP, and BSN on ActivityNet-1.3.}
\label{table:anet_prop}
\vspace{-0.1in}
\end{table}

Table~\ref{table:anet_prop} summarizes the performance comparisons on the validation and testing sets of ActivityNet-1.3. We compare our method with several other state-of-the-art methods in terms of AUC and AR@100. The results send a clear message that our method is more effective than all competitors. Specifically, our method improves AUC on the test set from previous performance record 66.26\% in the literature to 67.01\%.

Another desirable property of an action localization method is generating proposals for unseen action classes. We evaluate the generalization ability of our method on ActivityNet-1.3 validation set. Following the setting in~\cite{XiongZWLT17}, we choose 100 action classes (seen classes) videos from the training set to learn the model, and then evaluate the AUC / AR@100 for 100 seen and 100 unseen classes videos from the validation set respectively. Table~\ref{table:anet_gene} shows only a slight drop in AUC and AR@100 on the 100 unseen classes videos, which proves that our method has great generalization ability.

\begin{table}[t]
\centering
\begin{small}
\begin{tabular}{lcc}
  \toprule[0.8pt]
    {} & {AUC(val)} &  {AR@100(val)}  \\
  \midrule[0.5pt]
    {Seen (100 classes)} & 64.55	& 69.30	\\
    {Unseen (100 classes)} & 63.87 & 68.73 \\
  \midrule[0.5pt]
\end{tabular}
\end{small}
\caption{\small Evaluation of generalizing to unseen classes for our method on ActivityNet-1.3. }
\label{table:anet_gene}
\vspace{-0.1in}
\end{table}

\begin{figure}[t]
\begin{center}
\includegraphics[width=0.85\linewidth]{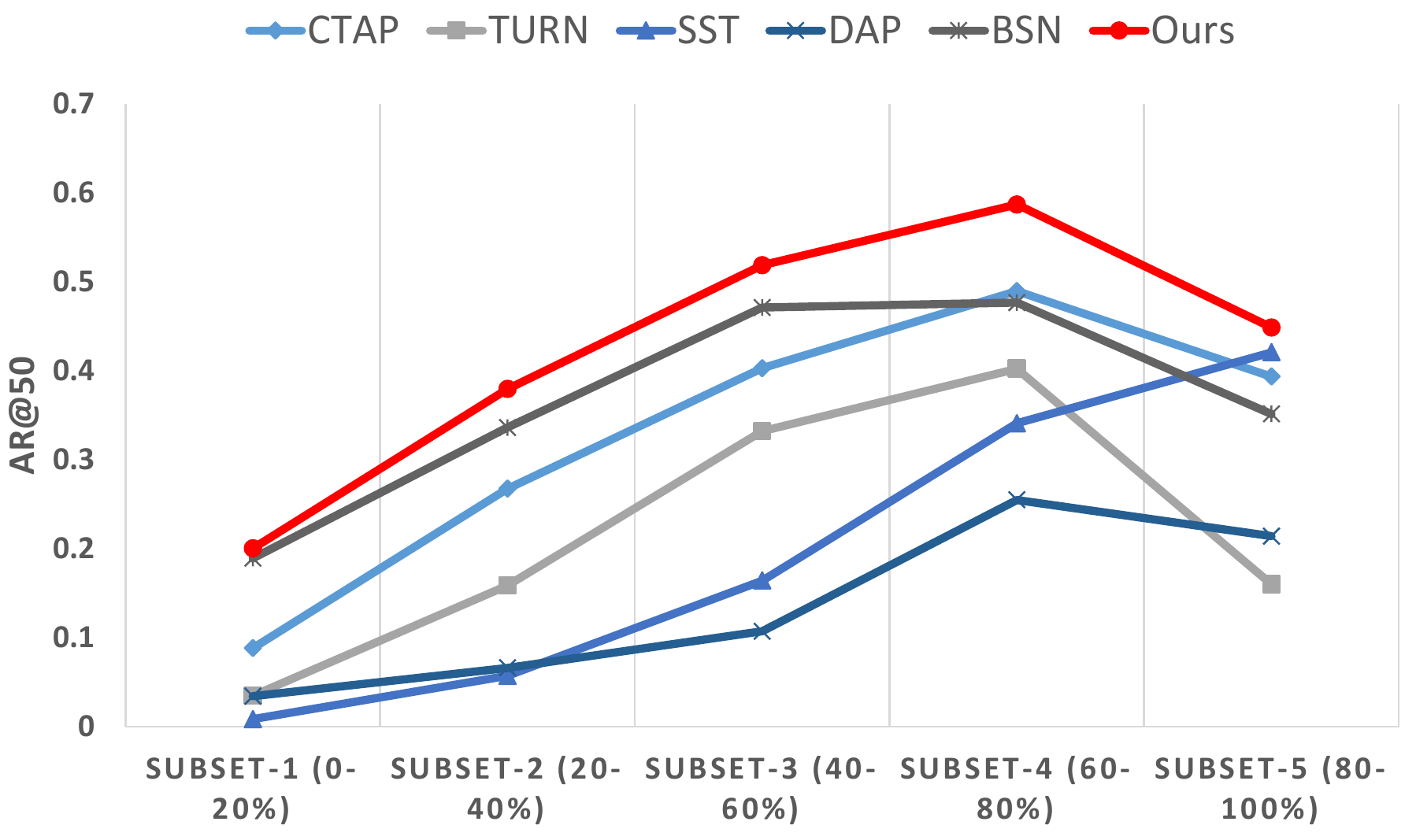}
\end{center}
   \caption{Comparing AR@50 on different-duration action instances on THUMOS14.}
\label{fig:duration}
\vspace{-0.15in}
\end{figure}

We also investigate separate performance on videos with different durations on THUMOS14. To this end, all action instances are cast into five sub-sets according to their durations. For example, ``subset-1 (0-20\%)' denotes the top-20\% instances with the smallest durations. Our proposed method strikes a clear superiority on all subsets.

\subsection{Ablative Study for Proposal Generation}

We investigate the contributions from several key components in the proposed TSA-Net. All experiments in this ablation study are performed on the THUMOS14 dataset with two-stream features.

\noindent \textbf{Effect of MDC Blocks}: To show the effectiveness of the MDC blocks in Figure~\ref{fig:network}, we compare two settings: 1) replace the MDC blocks of each branch with standard 1-D convolutions (TSA-Conv). For fairness, the kernel size of all 1-D convolution layers is set to 3 to ensure the same number of parameters with MDC blocks; 2) the original TSA-Net. Bayesian scores are used to rank proposals. Table~\ref{table:mtdcb_exp} shows the comparison of TSA-Conv with TSA-Net on THUMOS14 test set. It can be seen that the MDC blocks effectively boost the quality of the proposals via exploiting richer temporal contextual information.

\begin{table}[htbp]
\centering
\begin{small}
\begin{tabular}{cccc}
  \toprule[0.8pt]
  \multirow{2}{*}{Method} & \multicolumn{3}{c}{AR@AN} \\
  & {@50} & {@100} & {@200}  \\
  \midrule[0.5pt]
    {TSA-Conv} & 31.48 	& 40.83	& 47.78  \\
    {TSA-Net} & \textbf{40.83}	& \textbf{47.92}	 & \textbf{53.22} \\
  \bottomrule[0.8pt]
\end{tabular}
\end{small}
\caption{\small  Comparing TSA-Conv with TSA-Net on THUMOS14 in terms of AR@AN.}
\label{table:mtdcb_exp}
\end{table}

\noindent \textbf{Effect of Multi-branch Architecture}: We validate the design of multi-branch architecture in Figure~\ref{fig:network} by comparing four settings: 1) a single-branch network by setting MDC blocks' dilatation rates to (1,2,3) (denoted as \textbf{single-small}); 2) a single-branch network with MDC blocks with dilation rates (1,3,5) (denoted as \textbf{single-medium}); 3) a single-branch network with dilation rates (1,5,7) (let us call it \textbf{single-large}); 4) a multi-branch network which is composed of single-small, single-medium and single-large, referred to as \textbf{multi-branch}. Bayesian scores are adopted for ranking proposals. Table~\ref{table:multi_branch_exp} shows the comparison of all single-branch architectures with multi-branch architecture on THUMOS14 test set. It can be seen that the multi-branch network outperforms each single-branch network in terms of AR@AN. The results validate the effectiveness of the proposed multi-branch architecture and demonstrate that multi-scale temporal contextual information is important.

\begin{table}[htbp]
\centering
\begin{small}
\begin{tabular}{lccc}
  \toprule[0.8pt]
  \multirow{2}{*}{Method} & \multicolumn{3}{c}{{AR@AN}} \\
  & {@50} & {@100} & {@200}  \\
  \midrule[0.5pt]
    {single-small} & 37.72	& 45.85	& 52.03  \\
    {single-medium} & 37.77	& 45.01 & 50.38  \\
    {single-large} & 36.07	& 44.28	& 50.80 \\
  \midrule[0.5pt]
    {multi-branch} & \textbf{40.83}	& \textbf{47.92}	 & \textbf{53.22} \\
  \bottomrule[0.8pt]
\end{tabular}
\end{small}
\caption{Comparing single branch architectures with multi-branch architecture on THUMOS14 in terms of AR@AN.}
\label{table:multi_branch_exp}
\vspace{-0.15in}
\end{table}

\noindent \textbf{Effectiveness of Compatibility $\phi(c^{(s)}_i, c^{(e)}_i)$}: To evaluate the effectiveness of the learned pairwise compatibility function $\phi(c^{(s)}_i, c^{(e)}_i)$, we contrast the performance with or without the term $\phi(c^{(s)}_i, c^{(e)}_i)$ during scoring a proposal in Table \ref{table:TPRN_thumos}. These results clearly show that the proposed compatibility function improves the performance by non-trivial margins.

\begin{table}[htbp]
\centering
 \begin{small}
\begin{tabular}{cccc}
  \toprule[0.8pt]
  \multirow{2}{*}{Proposal Scoring} & \multicolumn{3}{c}{{AR@AN}} \\
  & {@50} & {@100} & {@200}  \\
  \midrule[0.5pt]
    {$P^{(s)}(c^{(s)}_i) \cdot P^{(e)}(c^{(e)}_i)$ in Eqn.~\ref{eqn:bayesian}} & 40.83	& 47.92	& 53.22  \\
    {$P \left( \overline{c^{(s)}_i c^{(e)}_i} \right)$ in Eqn.~\ref{eqn:bayesian}} & \textbf{42.83} & \textbf{49.61} & \textbf{54.52} \\
  \bottomrule[0.8pt]
\end{tabular}
 \end{small}
\caption{Effectiveness of compatibility function on THUMOS14 in terms of AR@AN.}
\label{table:TPRN_thumos}
\vspace{-0.15in}
\end{table}

\subsection{Experimental Analysis of Action Localization}

The evaluations performed on the testing set of THUMOS14 or ActivityNet-1.3 are shown in Table~\ref{table:thumos_det} or~\ref{table:anet_det} respectively. As previously stated, the development of action classifiers is relatively mature, and we thus directly re-use others' classification models for fairing contrasting different methods. The 5 baselines in topmost of Table~\ref{table:thumos_det} utilize their own classifiers. The rest baselines are either paired with the classifier in SCNN~\cite{ShouWC16}, Untrimmed Net~\cite{WangXLG17} or~\cite{zhao2017cuhk}. Most baselines adopt two-stream or C3D features while few use spatio-temporal features I3D~\cite{CarreiraZ17} (akin to the P3D features that we extract yet more computationally heavy), such as TAL-Net~\cite{ChaoVSRDS18}.

Table~\ref{table:thumos_det} exhibits that our proposed TSA-Net outperforms state-of-the-art action localization methods by significant margins when the IoU threshold varies from 0.3 to 0.7. Specifically, for mAP@0.5, our method significantly improves the performance from 36.9\% to 41.5\%, when both use the two-stream features. The performance of our method is further improved by using P3D features, and we achieved an mAP of 46.9\% when the IoU threshold is 0.5.

Similar trends are observed in Table~\ref{table:anet_det} on the benchmark ActivityNet-1.3. Note that the mAP scores on the testing set are obtained by submitting predictions to the competition server of ActivityNet-1.3, which only returns average mAP. As seen,  we improve the average mAP on the testing set from previous state-of-the-art 32.84\% (achieved by BSN) to 34.62\% using the same video features. On both benchmarks our method clearly re-calibrates the top performances.

\begin{table}[t]
\centering
\begin{scriptsize}
\begin{tabular}{lcccccc}
  \toprule[0.8pt]
  {Proposal Method} & {Classifier} & {0.7} & {0.6} & {0.5}  & {0.4} & {0.3} \\
  \midrule[0.5pt]
    {SCNN~\cite{ShouWC16}}		    & -- & 5.3	& 10.3	& 19	& 28.7	& 36.3 \\
    {CDC~\cite{ShouCZMC17}}		    & -- & 8.8	& 14.3	& 24.7	& 30.7	& 41.3 \\
    {R-C3D~\cite{XuDS17}}		    & -- & 9.3	& 19.1	& 28.9	& 35.6	& 44.8 \\
    {CBR~\cite{GaoYN17}}		        & -- & 9.9	& 19.1	& 31.0	& 41.3	& 50.1 \\
    {TAL-Net~\cite{ChaoVSRDS18} (I3D)} 	& -- & 20.8	& 33.8	& 42.8  & 48.5	& 53.2 \\
  \midrule[0.5pt]
    {SST~\cite{BuchESGN17}} & {SCNN-cls~\cite{ShouWC16}}& n/a	&n/a	&23.0	&n/a	&n/a  \\
    {TURN~\cite{GaoYSCN17}} & {SCNN-cls~\cite{ShouWC16}}& 7.7&14.6	&25.6	&33.2	&44.1  \\
    {BSN~\cite{LinZSWY18}} & {SCNN-cls~\cite{ShouWC16}}& 15.0	&22.4	&29.4	&36.6	&43.1  \\
    {CTAP~\cite{GaoCN18}} & {SCNN-cls~\cite{ShouWC16}}& n/a &n/a	&29.9	&n/a	&n/a  \\
    {Ours(Two-Stream)} & {SCNN-cls~\cite{ShouWC16}}& 17.1 &25.2	&33.1	&39.2	&43.7 \\
    {Ours(P3D)} & {SCNN-cls~\cite{ShouWC16}}& \textbf{19.4} & \textbf{27.8} & \textbf{36.6} & \textbf{43.7} & \textbf{48.3} \\
  \midrule[0.5pt]
    {SST~\cite{BuchESGN17}} & {UNet~\cite{WangXLG17}}& 4.7	&10.9	&20.0	&31.5	&41.2  \\
    {TURN~\cite{GaoYSCN17}} & {UNet~\cite{WangXLG17}}& 6.3	&14.1	&24.5	&35.3	&46.3  \\
    {BSN~\cite{LinZSWY18}} & {UNet~\cite{WangXLG17}}& 20.0	&28.4	&36.9	&45.0	&53.5  \\
    {Ours(Two-Stream)} & {UNet~\cite{WangXLG17}}& 21.7 &31.5	&41.5	&48.1	&53.2 \\
    {Ours(P3D)} & {UNet~\cite{WangXLG17}} & \textbf{25.2} & \textbf{36.1} & \textbf{46.9} & \textbf{55.9} & \textbf{61.2}  \\
  \bottomrule[0.8pt]
\end{tabular}
\end{scriptsize}
\caption{\small Action classification comparisons on THUMOS14 in terms of mAP@IoU. ``--" denotes that a method uses its own action classifier. ``n/a" denotes that the corresponding performance is not reported in the original literature.}
\label{table:thumos_det}
\vspace{-0.15in}
\end{table}

\begin{table}[h]
\centering
\begin{scriptsize}
\begin{tabular}{ll|cccc|c}
  \toprule[0.8pt]
  \multirow{2}{*}{Proposal Method} & \multirow{2}{*}{Classifier} & \multicolumn{4}{c|}{{Valid Set}} & {Test Set}\\
  & & {0.5} & {0.75} & {0.95} & {Average} & {Average} \\
  \midrule[0.5pt]
    {SCC~\cite{HeilbronBEG17}} & -- & 40.00	& 17.90	& 4.70 & 21.70 & 19.30  \\
    {CDC~\cite{ShouCZMC17}} & -- & 45.30 	& 26.00	& 0.20 & 23.80 & 22.90   \\
    {TCN~\cite{DaiSZDC17}} & -- & -	& -	& - & - & 23.58   \\
    {SSN~\cite{ZhaoXWWTL17}} & -- & 39.12	& 23.48	& 5.49 & 23.98 & 28.28   \\
    {TAL-Net~\cite{ChaoVSRDS18}} & -- & 38.23	& 18.30	& 1.30 & 20.22 & - \\
    {BSN~\cite{LinZSWY18}} & \cite{zhao2017cuhk} & 46.45	& 29.96	& 8.02 & 30.03 & 32.84 \\
  \midrule[0.5pt]
    {Ours} & \cite{zhao2017cuhk} & \textbf{48.71} & \textbf{31.97} & \textbf{8.97} & \textbf{31.90} & \textbf{34.62}  \\
  \bottomrule[0.8pt]
\end{tabular}
\end{scriptsize}
\caption{\small Action classification comparisons on the valid / test sets of ActivityNet-1.3.}
\label{table:anet_det}
\vspace{-0.15in}
\end{table}

\section{Conclusions}
\vspace{-0.in}

We propose TSA-Net for video action localization which effectively compiles temporal context from multi-scales. The core designs include a multi-branch stacked MDC blocks for temporal aggregation and a light-weight sub-network for regressing the proposal's Bayesian confidence. On large-scale benchmarks THUMOS14 and ActivityNet-1.3, the proposed TSA-Net outperforms competing methods and re-calibrate the state-of-the-art performances.

\clearpage

\end{document}